\documentclass{article} 
\usepackage{times}  
\usepackage{helvet}  
\usepackage{courier}  
\usepackage{graphicx} 
\usepackage{algorithm}
\usepackage{algorithmic}

\usepackage{amsmath}
\usepackage{amsthm}
\usepackage{booktabs}
\usepackage{amsmath,amsthm,amssymb}

\usepackage{newfloat}
\usepackage{listings}
\floatstyle{ruled}
\newfloat{listing}{tb}{lst}{}
\floatname{listing}{Listing}
\title{HydraGAN --  A Multi-head, Multi-objective Approach to Synthetic Data Generation}
\author{
    Chance DeSmet, Diane J. Cook
}
\usepackage{bibentry}

\begin{document}

\maketitle

\begin{abstract}
Synthetic data generation overcomes limitations of real-world machine learning. Traditional methods are valuable for augmenting costly datasets but only optimize one criterion: realism. In this paper, we tackle the problem of generating synthetic data that optimize multiple criteria. This goal is necessary when real data are replaced by synthetic for privacy preservation. We introduce HydraGAN, a new approach to synthetic data generation that introduces multiple generator and discriminator agents into the system. The multi-agent GAN optimizes the goal of privacy-preservation as well as data realism. To facilitate multi-agent training, we adapt game-theoretic principles to offer equilibrium guarantees. We observe that HydraGAN outperforms baseline methods for three datasets for multiple criteria of maximizing data realism, maximizing model accuracy, and minimizing re-identification risk. 
\end{abstract}

\section{Introduction}
\label{intro}
The growing availability of personal computational resources has resulted in an explosion in the amount of personal data that are collected and, occasionally, disseminated.  From banking details to recorded preferences and detailed behavior sensor readings, IoT devices have been the harbingers of a massive increase in the amount of insightful and sensitive individualized data.  Private data may not need to come from a source that is specifically focused on gathering information. For example, smart grid information may introduce privacy concerns, even though the system does not ostensibly exist to track users \cite{Chin2018ConsideringMeters, Zhou2019DifferentialData}.  Accompanying this rapid increase in big data production and consumption is the need to protect sensitive content from malicious adversaries.  Sharing data can support collaboration and promote new findings and advances, yet sharing sensitive data may violate privacy constraints and even endanger individuals whose personal information can be derived from the data  \cite{ElEmam2011AData,Winkler2004Re-identificationMicrodata}. 

To ensure that sensitive information is not leaked, many privacy-preserving data mining methods have been suggested in the literature \cite{Aggarwal2008AAlgorithms,Wagner2018TechnicalMetrics,Wang2019ResearchMining, Zhou2019DifferentialData}.  These methods all face a common privacy/utility trade-off. Specifically, an increase in privacy is accompanied by a commensurate decrease in data realism, and thus model accuracy \cite{Dong2018QuantifyingThings,Li2009OnPublishing,Snoke2018GeneralData}. In the most extreme case, privacy can be preserved by removing sensitive information or modifying these featurs beyond recognition. However, each modification may introduce changes and biases in the underlying concepts that can be learned. This negative correlation between data privacy and data utility may also be perceived as two ends of a spectrum. Therefore, the choice of Privacy-Preserving Data Mining (PPDM) method is often one that adapts best to the goals of the analysis task, such as the desired amount of privacy or the required utility of the predictive model.

To optimize the multiple competing goals of data realism, model integrity, and data privacy, we introduce HydraGAN, a multi-agent generative adversarial network (GAN) that game-theoretically optimizes multiple criteria to generate privacy-preserving, realistic synthetic data. We hypothesize that data generated by HydraGAN meet the multiple criteria that are required for privacy-preserving data mining. HydraGAN offers several novel contributions:
\begin{enumerate}
    \item HydraGAN introduces a privacy-preserving discriminator into a GAN framework.
	\item HydraGAN divides the data space among multiple generators to promote data diversity.
	\item HydraGAN ensures privacy by adversarially attempting to reidentify sensitive attributes. Instead of using a static preservation method such as noise infusion, our system trains a discriminator to reidentify sensitive attributes from the generated synthetic data. This discriminator dynamically improves its ability through training, simulating an actual `information warfare'' attack an adversary would make to access private information. This process conditions the generators to not provide data that are vulnerable to reidentification.
	\item HydraGAN is designed as a ``multi-headed'' GAN that combines any number of generators and discriminators. The method regulates the multi-agent training process using game-theoretic techniques.
\end{enumerate} 

In this paper, we introduce the notion of a multi-agent GAN and formally verify that the game-theoretic algorithm will reach an equilibrium between all contributing agents. We discuss how this method can be instantiated to perform privacy-preserving data mining. To evaluate algorithm performance, we compare HydraGAN with other reported methods on real and synthetic datasets using the multiple performance metrics of data utility, data privacy, and data realism.

\section{Related Work}
\label{related}

\subsection{Privacy Preserving Data Mining}
The goal of-privacy preserving data mining (PPDM) is to analyze sensitive data in a manner that does not reveal sensitive information.  PPDM techniques range from removing the sensitive information \cite{Chamikara2018EfficientMining} or adding noise to the data before analysis \cite{Rodriguez-Garcia2017AData} to generating entirely new data based upon the original \cite{Kurz2019TheSeries}. HydraGAN fits into this last category by generating synthetic data that both retain the privacy of sensitive information and are faithful to the original data.

\subsection{Differential Privacy}
Differential privacy is an increasingly common PPDM metric used to measure if the inclusion or removal of an individual's information would incur a privacy loss greater than a given bound, $\epsilon$ \cite{Dwork2006DifferentialPrivacy, Hilton2018DifferentialSurvey, Kifer2011NoPrivacy}.  The metric offers quantitative bounds and adapts to varied use cases  \cite{Asi2019ElementPrivacy,Cheu2019DistributedShuffling,Huang2020DP-ADMM:Privacy,Lecuyer2019CertifiedPrivacy}. Since its inception in 2006 \cite{Dwork2006DifferentialPrivacy}, differential privacy has rapidly increased in popularity and is now considered the 'de facto' privacy metric across a broad spectrum of domains \cite{Yang2020LocalSurvey}.   Liu et al. \cite{Liu2019DifferentialData} explore the use of differential privacy to quantify preservation of user identifiers in an eye-tracking study.  Their PPDM method added noise to the amount of time a person spent looking at each pixel on 'gaze maps.'  The amount of noise added to each gaze map ensured that adversaries would not be able to infer the form of a given individual's gaze map, even if an aggregated noise map was collected and the adversary held prior knowledge of other individuals.  Differential privacy may also be extended to meet increased privacy requirements. For example, local differential privacy provides a stronger guarantee than standard differential privacy, as it necessitates that an individual's data must be privatized before it is collected. To scale their work, Cormode et al. \cite{Cormode2019AnsweringPrivacy} created a distributed version of local differential privacy.  This extension afforded individuals the privacy guarantees of local differential privacy under more specific, and difficult to privatize, queries.  

\subsection{Reidentification Risk}

Because PPDM secures sensitive information, assessing effectiveness equates to measuring the risk of reidentifying sensitive information.   A primary type of reidentification attack is known as a 'linkage attack.' This type of attack attempts to combine multiple records for a single person to piece together sensitive information.  As an example, if an adversary had access to an individual's work schedule, along with their electric bill, the adversary could combine these insights to deduce when the person was away from their home and could subsequently cause harm to the person.  Over time, linkage attacks have become more sophisticated, using techniques such as probabilistic means to establish likely correlations between users \cite{Benitez2010EvaluatingRule,Malin2006Re-identificationRecords.,Sunteb1969ALinkage}.  HydraGAN protects against these linkage attacks by removing the original person from the data.  Any 'individual' present in the data generated by HydraGAN is in reality an amalgamation of the data from many different persons that trained the model, with additional privacy provided through its active reidentification discriminator.  Because of this process, HydraGAN-generated data would be difficult to use in a linkage attack.

\subsection{Generative Adversarial Networks}
\label{gansec}
In recent years, Generative Adversarial Networks have generated increasingly realistic synthetic data, evolving from relatively simple tasks to complex, multi-target generation.  Introduced by Goodfellow et al. \cite{Goodfellow2014GenerativeNets}, a GAN typically consists of two networks, a generator that proposes specific data configurations, and a discriminator that attempts to differentiate generated data from real samples.  As the generator learns how better to fool the discriminator, the discriminator correspondingly improves its perceptiveness at discovering subtle deviations from the original data.   Recent works have updated GAN design to include multiple generator or discriminator networks. For instance, CycleGAN \cite{Zhu2017UnpairedNetworks}, uses two discriminators and two generative mappings to learn how to translate images between different domains.  Similarly, Hardy et al. introduced MDGAN \cite{Hardy2019MD-GANDatasets}, that extends federated learning to a series of discriminators that represented by multiple distributed system.  This method allowed a single, central generator to efficiently learn from multiple distributed systems.  These GAN innovations set the stage for HydraGAN, a method that attempts to take advantage of the flexibility of multi-agent GANs to generate synthetic, private data.

\subsection{Synthetic Data Generation}
HydraGAN's ensures the privacy of sensitive data by generating synthetic data that obfuscates sensitive information while retaining the predictive concepts within the original data. Synthetic data generation techniques vary widely in form, including probabilistic approaches to simulating epidemiological information \cite{Eno2008GeneratingPatterns,Jalko2019Privacy-preservingModelling,Reiter2002SatisfyingSets}, deep methods such as autoencoders that create synthetic medical images \cite{TremblayTrainingRandomization,Xie2018DifferentiallyNetwork,Zhang2018StackedData}, and Markov models being that synthesize and predict network traffic \cite{Kordnoori2020TestingTraffic}.

Due to their strength in generating realistic data, GANs have recently become the de facto method for creating synthetic data. GAN-based synthetic data generation is primarily limited to image data. One influential work is med-GAN which translates medical images into new domains, assisting clinicians \cite{Armanious2020MedGAN:GANs}. Despite the resourceful data generation design, there is a potential for an adversary to discover the training examples used in the creation of synthetic data. To combat this attack, a new class of generator was introduced that attempts to secure the original data from malicious attack.
Two examples are PATE-GAN \cite{Jordon2019PATE-GaN:Guarantees} and PPGAN  \cite{Liu2020PPGAN:Network}. These methods, like HydraGAN, build on a Generative Adversarial Network (GAN) structure to generate privacy-preserving synthetic data.
PATE-GAN provides privacy by partitioning the original data set, training a set of discriminators on different data partitions, then noisily aggregating the results to be able to offer differential privacy guarantees to the resultant data \cite{Jordon2019PATE-GaN:Guarantees}.  PPGAN instead provides privacy by adding noise to the gradients of the discriminator, thus also offering a demonstrable amount of differential privacy to the generated data \cite{Liu2020PPGAN:Network}.

In contrast to these approaches, the goal of the HydraGAN algorithm is not to achieve differential privacy by adding uncertainty to the training process.  We do not build on such prior work because adding noise may degrade the performance of models that rely on data quality. We believe that dynamically learning to safeguard against possible attacks may be more protective than static, one-time noise addition methods.  To this end, HydraGAN learns to safeguard the data much the same way that a standard GAN learns to generate real data: by letting the generator iteratively compete against an adversary that wants to compromise the data. 

\section{Framework}
HydraGAN builds upon the Wasserstein GAN (WGAN) \cite{Arjovsky2017WassersteinNetworks}. Like a WGAN, HydraGAN contains a generator/discriminator pair in which the discriminator grades the realism of samples produced by the generator. From this starting point, we propose several new design elements to facilitate the optimization of multiple criteria. In this paper, we focus on the criteria of data realism and privacy preservation. This goal is accomplished through several changes to the traditional structure: the inclusion of a multi-headed generator, a re-identification discriminator, and a game-theoretic loss function to enable these components to interact appropriately. 

\begin{figure}
    \centering
    \includegraphics[scale=0.25]{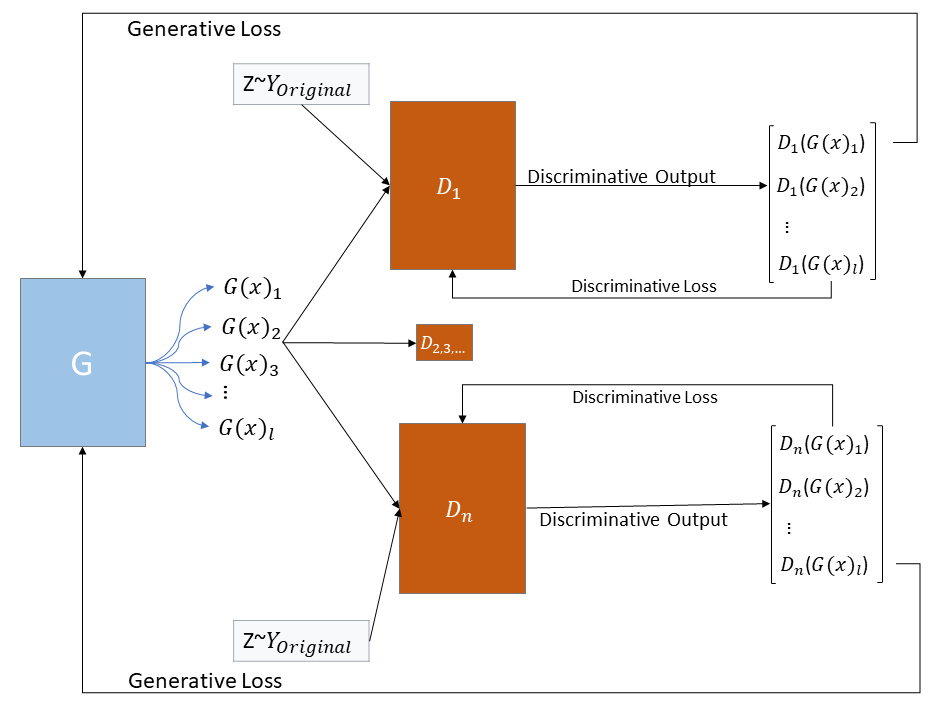}
    \caption{The HydraGAN multi-agent architecture.}
    \label{Network_figure}
\end{figure}

Figure \ref{Network_figure} illustrates the HydraGAN architecture. To prevent mode collapse, the generator contains multiple ``heads'', or generator functions, that operate on independent subsets of the real data. The multi-headed generator initially processes a noise vector and produces sets of synthetic samples, one set per generator head. Each set of samples is evaluated by a pair of discriminators, tasked with evaluating realism and re-identification. The discriminators process samples from each generator head separately and score each generated output independently.  After each training iteration, each learning agent (i.e., each generator head and each discriminator) is updated based on the loss from the previous iteration.  This update uses a custom loss function based on Equations \ref{D_cost} through \ref{equation_end} to synchronize training for all the HydraGAN agents.


\subsection{Multi-head Generator}
\label{gen_sec}
GANs are plagued by the issue of mode collapse \cite{Zhu2017UnpairedNetworks}, which increases the difficulty of training the agents.  Mode collapse often occurs when either the generator or discriminator diverges from its counterpart, resulting in the GAN producing unrealistic or repetitive synthetic samples \cite{Srivastava2017VEEGAN:Learning}. There are several common causes of mode collapse. A generator may  produce data that are so similar to the original that the discriminator cannot differentiate between the two, resulting in the generator solely producing samples within this small distribution.  On the discriminative side, a discriminator may learn to differentiate between the original and generated data to a degree that the generator is not able to fool the discriminator. In this case, the generator starts creating the same data continuously without any additional learning.

To lessen the impact of mode collapse , we partition the real data into clusters using k-means clustering. We select $k$ using the ``elbow'' technique \cite{Bholowalia2014EBK-Means:WSN}, increasing $k$ until the change in performance measure (e.g., the sum of squared distances between points and cluster centroids) converges.  

\begin{figure*}
    \begin{centering}
    \includegraphics[width=5.5in]{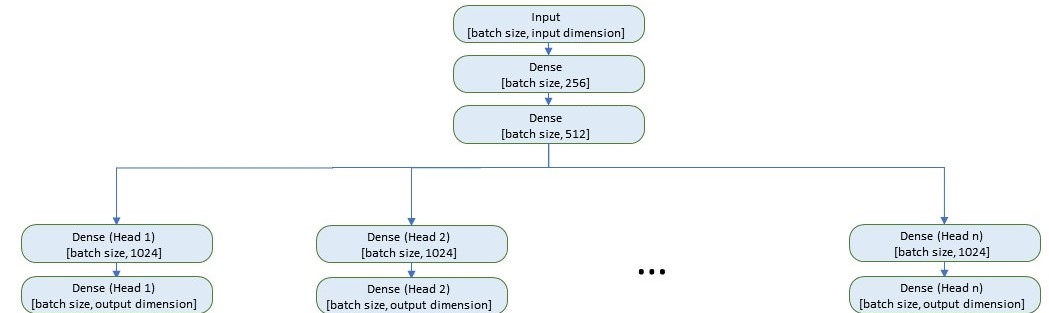}
    \caption{The network structure of the multi-head generator.}
    \label{Gen_table}
    \end{centering}
\end{figure*}

Next, we create a generator with multiple output ``heads''. Here, every generative head originates from a shared series of layers.  Figure \ref{Gen_table} contains the details of the generator construction.  This version of the generator contains  five heads, each outputting data described by fourteen features.  Each head is the same shape, and they all originate from a single layer in the generator, as shown in the figure. Each head is connected to an individual discriminator pair, and every discriminator/head pair is responsible for simulating one of the clusters.  This partitioning can also be valuable if different data subsets are accompanied by different privacy needs.  Through this construction, mode collapse is addressed in two ways. First, the simplification of the generated data distributions facilitates a more easily learned loss landscape for the generator and discriminator. Second, as each generative head is propagated from a shared set of layers, a generative head suffering from mode collapse may be fixed through updates to these initial layers originating from the other, better performing, heads.


\begin{figure}
    \centering
    \includegraphics[scale=0.3]{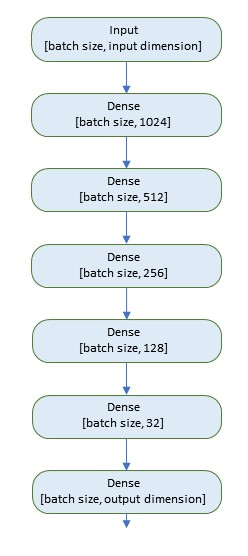}
    \caption{The network structure of the discriminators in HydraGAN.}
    \label{real_table}
\end{figure}

\subsection{Realism Discriminator}
HydraGAN's realism discriminator is tasked with gauging how faithfully the generated data emulate the real data. Figure \ref{real_table} illustrates this discriminator's structure. The network processes input data with fourteen features, convolves the input, and applies dense layers before a final score is produced. Convolutional layers are common for image GANs. We employ them in HydraGAN as well to maintain relationships between data features.  After training, this discriminator outputs a single variable, $r \in [-\inf,\inf]$, indicating the likelihood that the input data are real. As with other WGANs, gradient clipping is added to ensure that the network training occurs within a Lipshitz space \cite{Arjovsky2017WassersteinNetworks}.  

HydraGAN performance varies based on the activation functions that are employed.  We observed a faster training time when the activation function did not have a constant maximum, but yielded depreciating returns as the the input to this activation increased. As a result, we introduce the activation function shown in Equation \ref{logeq}. This operates like a Leaky ReLU with the nonlinear ``leaks'' occurring both below and above the range [-1  1]. 

\begin{equation}
\label{logeq}
F(x) = \frac{x}{|x|}*\log(|x| + 1)
\end{equation}   

\subsection{Re-Identification Discriminator}
In contrast to the realism discriminator, the re-identification discriminator's objective is not to discern real data from synthetic.  Instead, this discriminator grades the vulnerability of the generated data to re-identification.
To accomplish this goal, the discriminator actively attempts to identify the value of a sensitive feature given the non-sensitive features from the same data point.
Within HydraGAN, a quadratic loss function is used to incentivize the generator to produce samples that cause the prediction given by the re-identification to be different than what was actually output by the generator.  As the re-identification discriminator trains on real samples to learn how to re-identify the sensitive attribute, the generator will have to iteratively produce samples that do not lend themselves to an easy inference of the sensitive attribute (Equation \ref{quadloss}).  Additionally, in order to improve training speed, we do not activate the re-identification discriminator until the EM distance between the generated and real data is less than 0.3.

\begin{equation}
\label{quadloss}
F(y_{true},y_{pred}) = (|y_{true}-y_{pred}| - 0.25)^2
\end{equation}   

%

\subsection{Proof of Game-Theoretic Equilibrium}
\label{gameloss}
In the traditional GAN, the game-theoretic interplay is evident.  The generator and discriminator are bound together in a zero-sum game; thus, one cannot perform better without its counterpart performing worse.  The two networks form a minimax `game,' where a Nash equilibrium is reached when both networks are performing close to the same level and both have reached a local (or global) optimal state.  When transitioning to a multi-player GAN, this equilibrium is not guaranteed to exist.  In practice, this could be due to the generator successfully fooling the re-identifier while performing poorly against the realism discriminator.  Formulating the multi-objective equilibrium presents a challenge. When the generator adjusts its model to yield more realistic data, the resulting network may experience deteriorating re-identification performance.  However, this trend may reverse on subsequent training rounds.  Under these constraints, there is no guarantee that a convergence will ever occur, as the networks may oscillate between several different poorly-performing states.  We change the situation by modifying the WGAN loss function to also include the summed losses of the other HydraGAN agents.  This change links the networks into a collaborative partnership; an individual agent cannot perform poorly without its poor performance impacting the training of the other GAN agents.

We formalize our proposed approach by adapting the convergence equation introduced by Kodali et al.  \cite{Kodali2017OnRejected}. Here, we offer a proof that using the modified HydraGAN loss function, if the generator and a discriminator reach an equilibrium, that point also represents an equilibrium for the rest of the discriminators.

In this proof, we first define the cost functions for a standard GAN and describe how those cost functions establish an equilibrium between the single generator / discriminator pair.  We then introduce our new cost function. Using the new cost function, we verify that when an equilibrium is reached for a single generator-discriminator pair, it follows that an equilibrium is reached for all of the HydraGAN agent pairs. 

We first restate the traditional GAN cost function, shown in Equations \ref{G_cost} and \ref{D_cost}. In these equations, $J_{n}(\phi, \theta)$ represents the cost function for a network $n$ with parameters $\phi$ and $\theta$. These parameters represent the current weights of the generator and discriminator, respectively. The right hand side of the equations follow the notation introduced by Goodfellow et al. \cite{Goodfellow2014GenerativeNets} for the generator and discriminator components of the loss function. In these equations,
$E_{X}$ draws a sample from the real data distribution and $E_{z}$ draws a sample from some input noise.  Next, $D_{\phi}(x)$ represents the performance of the discriminator given samples $x$ with weights $\theta$, and $G_{\theta}(z)$ represents the generator's output based on input $z$ using weights $\theta$. The goal of the original two-player GAN is to find parameters $\theta$ that minimize the log probability of $1-D(G(z))$. The term $D(G(z))$ represents the probability that generated data $G(z)$ is real. If the discriminator correctly classifies a fake input, then $D(G(z))=0$.

%


\begin{proof}
      \small
      \begin{equation}
        J_{G}(\phi,\theta) := E_{X~p\in{real}} \log D_{\theta}(x) + E_{z} \log(1-D_{\theta}(G_{\phi}(z)))
        \label{G_cost}
        \end{equation}
        \begin{equation}
        J_{D}(\phi,\theta) := -E_{X~p\in{real}} \log D_{\theta}(x) - E_{z} \log(1-D_{\theta}(G_{\phi}(z))) 
        \label{D_cost}
        \end{equation}
        \normalsize

We first define the cost functions for both the generator and the discriminator in a two-agent system.  Equation \ref{D_cost} represents the cost from the perspective of the discriminator, evaluating how well it can identify if the input values $x$ are drawn from the real data distribution or if they are synthetic examples provided by the generator.  Equation \ref{G_cost} describes the cost function in terms of the generator. This cost function considers the generator's performance in fooling the discriminator into labeling its output as ``real''.

Equations \ref{G_equib} and \ref{D_equib} introduce the equilibrium equations from Kodali et al. \cite{Kodali2017OnRejected}.  These equations use the cost function $J_{n}(\phi, \theta)$ to describe an equilibrium between the generator and the discriminator.  What these equations show is that for the parameters $\phi$ and $\theta$, a change of these parameters within some bound $\gamma$ will not result in a superior cost function performance for either network.  The term $'$ represents the current parameterization of the network, and $*$ represents a new parameterization within some maximum distance $\gamma$ away from $'$.  An $\epsilon$ term is included to provide an error bound.  Equations \ref{G_equib} and \ref{D_equib} thus represent an equilibrium such that all of the weights in the system are in a position where any change would result in diminished performance.
    \begin{equation}
    \forall \phi',||\phi'-\phi^{*} || \leq \gamma : J(\phi^{*},\theta^{*}) \leq J(\phi',\theta^{*}) + \epsilon 
    \label{G_equib}
     \end{equation}
     \begin{equation}
    \forall \theta',||\theta'-\theta^{*} || \leq \gamma : J(\phi^{*},\theta^{*}) \geq J(\phi^{*},\theta') - \epsilon 
    \label{D_equib}
    \end{equation}
In HydraGAN, multiple discriminators are present, each with its own objective.  To accommodate these various, sometimes competing goals, we introduce a new cost function in Equation \ref{add_lambda}.  In this example, the cost function references particular discriminator $j$.  Additionally, this function includes the individual cost of the remaining discriminators with the addition of a $\lambda$ term that sums this cost of all discriminators other than $j$. In the multi-agent formulation, calculating the loss of an agent must take into account the rest of the HydraGAN discriminators, essentially combining the losses.  We will now verify that using this new cost function  will allow us to arbitrarily choose an equilibrium for a specific generator/discriminator pair. By rearranging the terms in Equation \ref{pull_k},
we will demonstrate that this is an equilibrium point for all discriminators.
   \begin{equation}
   \begin{split}
    J_{D}(\phi,\theta_{j},\lambda) := E_{X~p\in{i}} \log(D_{\theta_{j}}(x)) + E_{z} \log(1-D_{\theta_{j}}(G_{\phi}(z)))\\ 
    +\sum_{i \in \lambda|i\neq j}{E_{X~p\in{\lambda_{i}}} \log(D_{\theta_{i}}(x)) + E_{z} \log(1-D_{\theta_{i}}(G_{\phi}(z)))} 
    \end{split}
    \label{add_lambda}
   \end{equation}
    
Equation \ref{new_equib} expresses the new equilibrium equation. In this question, $\lambda$ represents the rest of the cost functions present in HydraGAN.
    
   \begin{equation}
     \forall \theta',||\theta'-\theta^{*} || \leq \gamma : J(\phi^{*},\theta^{*},\lambda) \geq J(\phi^{*},\theta',\lambda) - \epsilon 
     \label{new_equib}
    \end{equation}
     We now label $\theta$ in terms with our selected discriminator $j$, as shown in Equation \ref{new_thet}:
   \begin{equation}
    \forall \theta',||\theta'-\theta^{*} || \leq \gamma : 
    J(\phi^{*},\theta^{*}_{j},\lambda) \geq J(\phi^{*},\theta_{j}',\lambda) - \epsilon \\
    \label{new_thet}
    \end{equation}
   Next, we expand the new cost functions found in Equation \ref{new_thet}, allowing us to view the equilibrium as a composite of the inner terms.
   \begin{equation}
   \begin{split}
    := E_{X~p\in{j}} \log(D_{\theta_{j}^{*}}(x)) +    E_{z}\log(1-D_{\theta_{j}^{*}}(G_{\phi^{*}}(z))) +\\
    \sum_{i \in \lambda|i\neq j}{E_{X~p\in{\lambda_{i}}} \log(D_{\theta_{i}^{*}}(x)) + E_{z} \log(1-D_{\theta_{i}^{*}}(G_{\phi^{*}}(z)))} \\
    \geq  E_{X~p\in{j}} \log(D_{\theta_{j}^{`}}(x)) + 
    E_{z} \log(1-D_{\theta_{j}^{`}}(G_{\phi}(z))) + \\
    \sum_{i \in \lambda|i\neq j}{E_{X~p\in{\lambda_{i}}} \log(D_{\theta_{i}^{`}}(x)) + E_{z} \log(1-D_{\theta_{i}^{`}}(G_{\phi}(z)))} - \epsilon 
    \end{split}
    \end{equation}
    
    We arbitrarily choose an additional discriminator $k$, and express the cost function associated for discriminator $k$, along with the cost for discriminator $j$ and the summation costs for the remaining discriminators.
    \begin{equation}
    \begin{split}
    :=E_{X~p\in{j}} \log(D_{\theta_{j}^{*}}(x)) +    E_{z}\log(1-D_{\theta_{j}^{*}}(G_{\phi^{*}}(z))) + \\
    E_{X~p\in{k}} \log(D_{\theta_{k}^{*}}(x)) +    E_{z}\log(1-D_{\theta_{k}^{*}}(G_{\phi^{*}}(z))) +\\
    \sum_{i \in \lambda|i\neq j,k}{E_{X~p\in{\lambda_{i}}} \log(D_{\theta_{i}^{*}}(x)) + E_{z} \log(1-D_{\theta_{i}^{*}}(G_{\phi^{*}}(z)))} \\
    \geq  E_{X~p\in{j}} \log(D_{\theta_{j}^{`}}(x)) +  E_{z}\log(1-D_{\theta_{j}^{`}}(G_{\phi}(z))) +\\
    E_{X~p\in{k}} \log(D_{\theta_{k}^{`}}(x)) +  E_{z}\log(1-D_{\theta_{k}^{`}}(G_{\phi}(z))) +\\
    \sum_{i \in \lambda|i\neq j,k}{E_{X~p\in{\lambda_{i}}} \log(D_{\theta_{i}^{`}}(x)) + E_{z} \log(1-D_{\theta_{i}^{`}}(G_{\phi}(z)))} - \epsilon \\
    \label{pull_k}
    \end{split}
    \end{equation}
    We now integrate the original discriminator $j$ into the $\lambda$ term representing the rest of the HydraGAN discriminators.  This is possible because all discriminators use the same cost function, as well as the same maximum step size.  This addition is a critical step, demonstrating that the $\lambda$ term representing the rest of the discriminator's loss functions allows us to ensure that any equilibrium point is shared between all discriminators.
   \begin{equation}
   \begin{split}
    := E_{X~p\in{k}} \log(D_{\theta_{k}^{*}}(x)) +    E_{z}\log(1-D_{\theta_{k}^{*}}(G_{\phi^{*}}(z))) +\\
    \sum_{i \in \lambda|i\neq k}{E_{X~p\in{\lambda_{i}}} \log(D_{\theta_{i}^{*}}(x)) + E_{z} \log(1-D_{\theta_{i}^{*}}(G_{\phi^{*}}(z)))} \\
    \geq  E_{X~p\in{k}} \log(D_{\theta_{k}^{`}}(x)) + 
    E_{z} \log(1-D_{\theta_{k}^{`}}(G_{\phi}(z))) +\\
    \sum_{i \in \lambda|i\neq k}{E_{X~p\in{\lambda_{i}}} \log(D_{\theta_{i}^{`}}(x)) + E_{z} \log(1-D_{\theta_{i}^{`}}(G_{\phi}(z)))} - \epsilon \\
    \end{split}
    \end{equation}
    
   \begin{equation}
    \forall \theta',||\theta'-\theta^{*} || \leq \gamma : J(\phi^{*},\theta_{k}^{*},\lambda) \geq J(\phi^{*},\theta_{k}',\lambda) - \epsilon \qedhere
    \label{equation_end}
    \end{equation}
Equation \ref{equation_end} is a transformed version of the original cost function \ref{new_thet}, replacing discriminator $j$ with $k$.  Following this process, we see that any equilibrium point between the generator and a given discriminator is an equilibrium point for every other discriminator as well. This result motivates the modification of the standard Wasserstein loss function for each discriminator into a loss function that also accounts for the other discriminator in HydraGAN, allowing for a more equitable training process that is not dominated by either the realism or re-identification discriminator.
   \begin{align*}
  \end{align*}
\end{proof}

\section{Experimental Results}

Our goal is to create synthetic data that satisfy multiple objectives. In this paper, our objectives are 1) data realism, 2) performance of a predictive model trained on the data, and 3) privacy preservation. In our work, we view privacy preservation as obscuring the values of sensitive features. If accessed either unintentionally or maliciously by an external source, sensitive features may cause damage to one or more parties. To evaluate the utility of HydraGAN-generated data, we compare its performance along these multiple dimensions with baseline synthetic data generation methods.  The first of the two hyperparameters tuned during the implementation and testing of this mode were the learning rates for both the RMSprop and Adam optimizers, which ranged during testing from [0.0000001, 0.1], with best performance at 0.0002. The second parameter was the clip constraint of the weight clipping, ranging in tests from [0.001, 0.5], with best performance at 0.05 

We validate the methods on three datasets. The first real-world dataset is the UCI Heart Disease dataset \cite{Janosi1988HeartSet}. In this data, each sample contains 13 physiological characteristics, including an attribute indicating whether the specific individual was diagnosed with heart disease. The second dataset contains 53 features extracted from ambient sensor data readings collected in smart homes  \cite{Schmitter-Edgecombe2012NaturalisticTask}. Sensor data were collected while subjects performed scripted activities in a smart home testbed. The testbed contains ambient sensors that monitor motion (with a passive infrared motion detector), ambient light and temperature, and door usage (with magnetic door closure sensors). In these two datasets relating to human participants, the sensitive value we will use will is participant age, as previous studies reveal this is a feature that is vulnerable to a re-identification attack \cite{Na2018FeasibilityLearning}.  The third dataset represents electrical power consumption, reporting load, reaction time, and power balance under various conditions, from a work by Arzamasov et al.  \cite{Arzamasov2018TowardsStability}  We treat the ``stability'' variable as sensitive for this analysis. 

\begin{figure}
    \centering
    \includegraphics[scale=0.27]{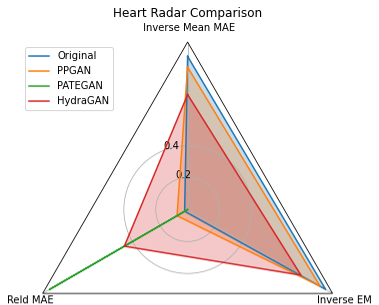}
    \caption{Performance comparison for UCI heart dataset.}
    \label{Heart_radar}
\end{figure}

The metrics of realism and privacy often compete; thus, we examine performance separately along these dimensions. To evaluate data realism, we will calculate the inverse of the Earth Mover's (EM) distance between the original and generated data (Inverse EM) \cite{Rubner2000TheRetrieval}.  To evaluate a model's predictive performance trained on the data, we compute average predictive error across three classifiers (support vector, K nearest neighbors, and decision tree regressor) trained on the data and trying to predict the values of non-sensitive features in the real data (Inverse Mean MAE).  To evaluate privacy preservation, we compute the average error of the same classifiers in predicting the sensitive attribute (ReID MAE).  A generator that performs well on these metrics will produce data that are of a similar form to the original, preserve the predictive utility of the real data, and safeguard the sensitive attribute.

\begin{figure}
    \centering
    \includegraphics[scale=0.27]{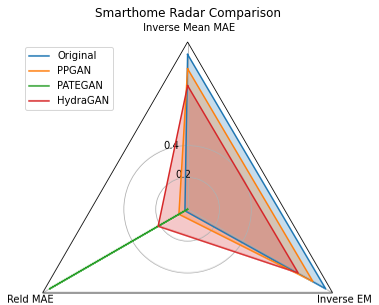}
    \caption{Performance comparison for smart home dataset.}
    \label{Group_radar}
\end{figure}

To visualize the performance of these models for multiple objectives, we use a radar plot highlighting the three performance dimensions. The goal is to maximize the value along each dimension. In addition, we will compute the correlation between the sensitive attribute and the remaining features. Paired with a visualization of the generated data, we can observe the impact each feature has on the re-identification of sensitive attributes.  

\begin{figure}
    \centering
    \includegraphics[scale=0.27]{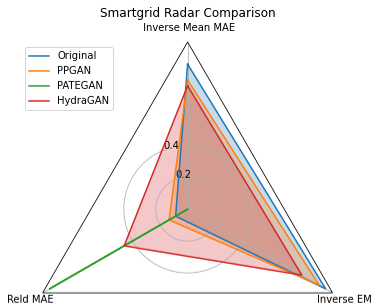}
    \caption{Performance comparison for smart grid dataset.}
    \label{Grid_radar}
\end{figure}

From the radar charts, we see that HydraGAN is more adaptive to  multiple objectives. In particular, HydraGAN attains greater realism than PATEGAN and provides greater privacy preservation than PPGAN. PPGAN focuses its effort more on realism than privacy preservation, yet HydraGAN's realism is comparable based on model MAE as well as EM distance, while providing a much greater level of privacy to the sensitive attributes of age (Figures \ref{Heart_radar} and \ref{Group_radar}) and power consumption (Figure \ref{Grid_radar}). As would be expected, as the method offers greater privacy, the data become less consistent with the original.  This corresponds with the well-known PPDM concept of the utility/privacy trade-off \cite{Dong2018QuantifyingThings}, where increasing the amount of privacy provided to participants in some data will always result in a cost to the realism of that data.

\begin{figure}
    \centering
    \includegraphics[scale=0.32]{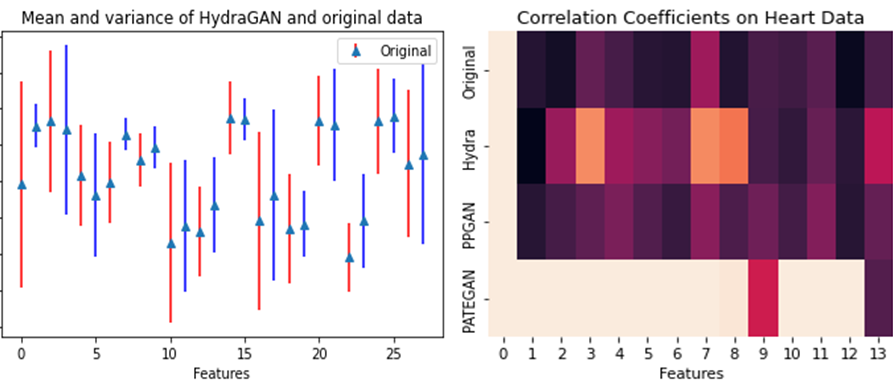}
    \caption{(left) Comparison of real and HydraGAN-generated features for the smart grid dataset; (right) Correlation coefficients between each feature and sensitive attribute \#0 for the four generation methods.}
    \label{corr}
\end{figure}

Next, we plot statistics of HydraGAN-generated data and the original UCI heart data. As Figure \ref{corr} (left) illustrates, the mean and standard deviation of most features align well. To understand why not all features align, we also plot the absolute value of the correlation coefficients between feature (numbered randomly in Figure \ref{corr}) and the sensitive attribute (feature 0 in the figure). As the figure illustrates, the stronger the correlation between a feature and the sensitive attribute, the greater is the difference between statistics for the original and synthetic feature values. While this difference can be accomplished by infusing noise, the radar charts indicate the dramatic detrimental impact this technique has on data realism and model predictability. Looking at Figure \ref{corr} (right), we further note that PPGAN preserves the relationship between the sensitive and non-sensitive features very well, while HydraGAN preserves some relationships and not others.  This may be due to HydraGAN's re-identification discriminator obscuring re-identification of the sensitive feature by blurring the relationships for highly-correlated features.

\begin{table}
  \begin{center}
    \caption{Comparison of original data and three privacy-preserving methods for the metrics of realism (Inverse EM), model predictability (Inverse Model MAE), and re-identification ( ReID MAE).}
    \label{tab:heart_samp}
    \begin{tabular}{lccc} 
      \hline\textbf{Method} & \text{Inverse} & \text{Inverse} & \text{ReID}\\
      & \text{EM} & \text{Model MAE} & \text{ MAE} \\
      \hline \hline
      \multicolumn{4}{c}{$Heart$}  \\
      Original & 1.00 & 0.96 & 0.02\\
      HydraGAN & 0.82 & 0.72 & 0.46\\
      PPGAN & 0.96 & 0.89 & 0.08\\
      PATEGAN & 0.00 & 0.00 & 1.00\\
      \hline
      \multicolumn{4}{c}{$Smart home$}  \\
      Original & 1.00 & 0.97 & 0.02\\
      HydraGAN & 0.80 & 0.78 & 0.21\\
      PPGAN & 0.91 & 0.88 & 0.06\\
      PATEGAN & 0.00 & 0.00 & 1.00\\
      \hline
      
      \multicolumn{4}{c}{$Smart grid$}  \\
      Original & 1.00 & 0.91 & 0.09\\
      HydraGAN & 0.83 & 0.77 & 0.46\\
      PPGAN & 0.95 & 0.81 & 0.13\\
      PATEGAN & 0.00 & 0.00 & 1.00\\
      \hline
    \end{tabular}
  \end{center}
\end{table}



\section{Conclusions}
We introduce a novel multi-objective synthetic data generator, called HydraGAN,.  This multi-agent GAN balances a need for realistic data with preserving private information. We validated the ability of HydraGAN to achieve a system equilibrium if one exists. Additionally, experiments real-world datasets indicate that HydraGAN can optimize these multiple, competing performance criteria, resulting in synthetic data that are realistic, retain the predictive performance of real data, and maintain the privacy of sensitive attributes. Further work will incorporate and evaluate additional discriminators.  Future work will also investigate refinements of the clustering method for promoting data diversity.


\bibliographystyle{abbrv}
\bibliography{main}

\end{document}